\documentclass[letterpaper, 10 pt, conference]{ieeeconf}  

\IEEEoverridecommandlockouts                              

\usepackage{graphicx} 
\usepackage{times} 
\usepackage{amsmath} 
\usepackage{amssymb}  
\usepackage{bbold}
\usepackage{url}
\usepackage{cite}
\usepackage{enumerate}
\usepackage{booktabs}
\usepackage{tabularx}
\usepackage{multirow}
\usepackage{pifont}
\usepackage[table,xcdraw]{xcolor}
\usepackage{xspace}
\usepackage{float} 
\usepackage[normalem]{ulem}
\useunder{\uline}{\ul}{}
\makeatletter
\let\NAT@parse\undefined
\makeatother
\usepackage[pagebackref=false,breaklinks=true,letterpaper=true,colorlinks,bookmarks=false]{hyperref}

\usepackage{colortbl}

\definecolor{baselinecolor}{gray}{.9}

\newcommand{\modelname}{LatentDriver}
\definecolor{actioncolor}{HTML}{F7931E}
\definecolor{statecolor}{HTML}{819274}

\newlength{\savedwidth}
\newcommand{\whline}[1]{\noalign{\global\savedwidth\arrayrulewidth \global\arrayrulewidth #1}%
\hline \noalign{\global\arrayrulewidth\savedwidth}}

\title{\LARGE \bf
Learning Multiple Probabilistic Decisions from Latent World Model in Autonomous Driving
}

\author{Lingyu Xiao$^{1,2*}$, Jiang-Jiang Liu$^{2\dag}$, Sen Yang$^{2}$, Xiaofan Li$^{2}$, Xiaoqing Ye$^{2}$, Wankou Yang$^{1\ddag}$ \\ and Jingdong Wang$^{2}$
\thanks{$^{*}$ Work done during an internship at Baidu.}
\thanks{$^{\dag}$  Project lead, $^{\ddag}$ Corresponding author. }
\thanks{$^{1}$Authors are with School of Automation, Southeast University, Nanjing, China. \texttt{\{lyhsiao,wkyang\}@seu.edu.cn}}%
\thanks{$^{2}$Authors are with Baidu Inc., Shanghai, China.}%
}

\begin{document}

\maketitle
\thispagestyle{empty}
\pagestyle{empty}

\begin{abstract}
The autoregressive world model exhibits robust generalization capabilities in vectorized scene understanding but encounters difficulties in deriving actions due to insufficient uncertainty modeling and self-delusion.
In this paper, we explore the feasibility of deriving decisions from an autoregressive world model by addressing these challenges through the formulation of multiple probabilistic hypotheses. 
We propose \emph{\modelname{}}, a framework models 
the environment's next states and the ego vehicle's possible actions 
as a mixture distribution, from which a deterministic control signal is then derived. 
By incorporating mixture modeling, the stochastic nature of decision-making is captured. Additionally, the self-delusion problem is mitigated by providing intermediate actions sampled from a distribution to the world model.
Experimental results on the recently released \textit{close-loop} benchmark Waymax demonstrate that \modelname{} surpasses state-of-the-art reinforcement learning and imitation learning methods, achieving expert-level performance. 
The code and models will be made available at \url{https://github.com/Sephirex-X/LatentDriver}.

\end{abstract}

\section{Introduction}

Motion planning is a fundamental task in autonomous driving systems. Recently, with the introduction of several real-world data-driven benchmarks~\cite{caesar2021nuplan,gulino2024waymax}, learning based planning methods have garnered significant attention from both industry and academia. However, navigating through various unfamiliar driving scenarios based on the vehicle's current observations remains extremely challenging. This difficulty arises from the complexity involved in understanding interactions among traffic participants and the unstructured nature of road environments. Most critically, it involves deriving appropriate actions from these observations.

\begin{figure}
    \centering
    \includegraphics[width=1\linewidth]{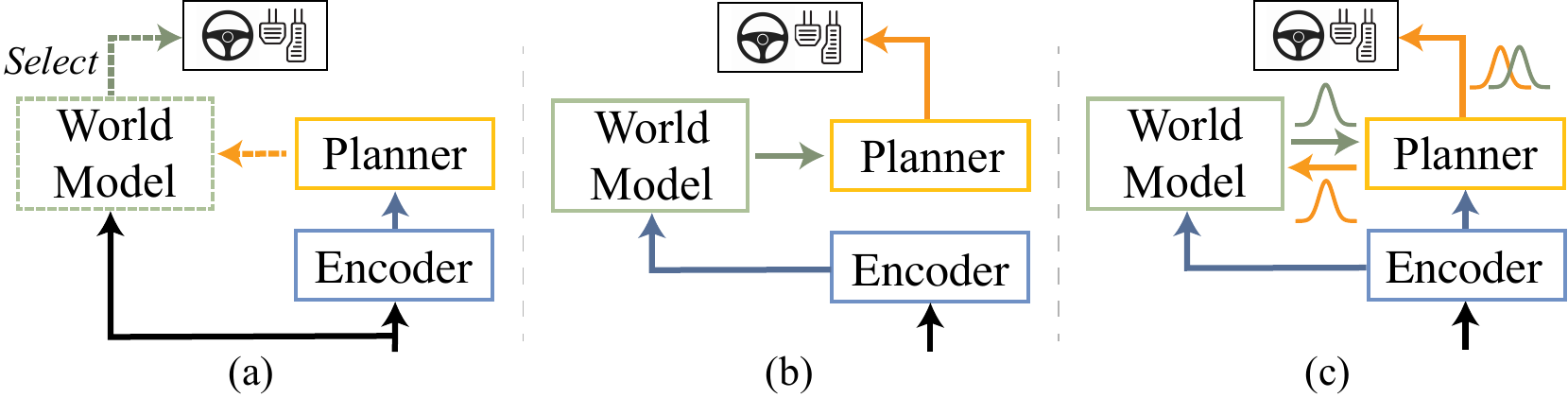}
    \caption{Different designs of world model integration. Dashed arrows indicate the absence of gradient. (a) Treats the world model as a realistic simulator and selects the best action from multiple planners (actions). (b) Directly derives actions from the world model's latent space. (c) Our method models the environment’s next states and the ego vehicle’s next possible actions as a mixture distribution and derives the ultimate action from it.}
    \label{fig.teaser}
    \vspace{-5mm}
\end{figure}

To understand the environment, early methods~\cite{pdm,renz2022plant} typically modeled current dynamics using frameworks like PointNet~\cite{qi2017pointnet} or BERT~\cite{devlin2018bert}. Recent works~\cite{cheng2024rethinking,cheng2024pluto,gulino2024waymax,gameformer} have designed encoders based on successful implementations from motion forecasting~\cite{nayakanti2023wayformer,cheng2023forecast}. Despite their effectiveness, performance in out-of-distribution scenes remains suboptimal. In response,~\cite{hu2024solving} developed an autoregressive world model with strong generalization capabilities for environmental understanding. As shown in Fig.~\ref{fig.teaser}(a), this model integrates with various planners, functioning as an interactive simulator that generates scores for each planner and selects the best one. However, performance is limited by the world model's imperfections and insufficient sparse signals for the planner. Additionally, the high training cost for a sufficiently accurate world model poses challenges. This raises the question: How can we leverage the world model's knowledge to aid planner learning at minimal cost?

One potential solution is to implicitly transfer knowledge from the world model to the planner and optimize them jointly~\cite{drivedreamer,mils,li2024enhancing}, as demonstrated in Fig.~\ref{fig.teaser}(b). However, these approaches fall short of fully utilizing the autoregressive model's potential. The first issue is the incomplete consideration of uncertainty, particularly regarding the ego vehicle's actions when interacting with the environment. Driving scenes are inherently stochastic, and decision-making should not be considered a single-modality problem. Multiple valid options may exist, with each option representing a different mode of the distribution. Another challenge is self-delusion. Learning in the autoregressive world model involves sequence prediction~\cite{radford2019language}, with agent interactions framed as cascading conditional distributions of actions. The planner, however, must respond based on current observations rather than historical actions, exacerbating the ‘copycat’ phenomenon~\cite{ortega2021shaking,cheng2024rethinking} in imitation learning planners.

To address the above issues, we propose \emph{\modelname{}} with a key insight: it hypothesize the distribution for actions and states is multi-probabilistic, as well as their combination. As illustrated in Fig.~\ref{fig.teaser}(c), the interaction between the world model and the planner is bi-directional and fully stochastic, with the final action derived from their mixture distribution.
Specifically, we introduce the Multiple Probabilistic Planner (MPP), which models the ego vehicle's actions as a stochastic process through a mixture Gaussian distribution~\cite{chai2019multipath,shi2022motion}. The MPP is structured in a multi-layer transformer style, with each layer refining action distributions based on the Latent World Model's (LWM) output. Therefore, it naturally capturing the stochastic actions of the ego vehicle. To mitigate the self-delusion problem during joint optimization, actions sampled from an intermediate layer of the MPP serve as an estimation of real actions, reducing the reliance on historical actions for final decisions.
Extensive experiments on Waymax~\cite{gulino2024waymax} demonstrate expert-level performance in evaluations against both non-reactive and reactive agents.

Our contributions are summarized as follows:

\begin{itemize}
\item Modeling the environment's next states and ego's next possible actions as a mixture distribution, better fitting the stochastic nature of decision-making.
\item A model that unifies the learning of autoregressive world model and ego planning without self-delusion.
\item Demonstration of expert-level performance in \textit{close-loop} simulations using Waymax, against both non-reactive and reactive agents.
\end{itemize}

\section{Related works}
\subsection{World Model for Planning in Autonomous Driving}
The world model aims to capture the transition dynamics of the environment using data. Two primary approaches exist for integrating world models into planning. The first treats the world model as an accurate simulator~\cite{wang2024driving,hu2024solving}, where actions are selected based on the lowest cost through simulation. This method focuses on maximizing world model's precision; for example, Drive-WM~\cite{wang2024driving} employs a diffusion model~\cite{stablediffusion} to generate action-conditional realistic images, while GUMP~\cite{hu2024solving} uses a GPT-style autoregressive model to learn dynamics from vectorized inputs. The effectiveness of this approach hinges on the simulator's accuracy and the associated training cost is significant. The second approach~\cite{drivedreamer,mils,ADriver-I,li2024enhancing}, treats world model learning as an auxiliary task, generating actions directly from the image feature space. DriveDreamer~\cite{drivedreamer} and MILE~\cite{mils} utilize VAE~\cite{kingma2013auto} and LSTM~\cite{lstm} to model transition dynamics, while ADriver-I~\cite{ADriver-I} employs LLaVA~\cite{llava}. However, these methods have not been validated in complex real-world \textit{closed-loop} evaluations and are unsuitable for vectorized observation spaces.

\subsection{Imitation Learning Based Planner}

Imitation learning based planners can be categorized by their observation space into end-to-end~\cite{hu2023planning,jiang2023vad,chen2022learning} and mid-to-end methods (also known as mid-to-mid). End-to-end approaches directly learn driving policies from raw sensor data, while mid-to-end methods rely on post-perception outputs. Our method falls into the latter category. Early works~\cite{vitelli2022safetynet,scheel2022urban,pini2023safe} validated their algorithms in real-world settings due to the absence of standardized benchmarks. 
More recent works, including PlanT~\cite{renz2022plant} and Carformer~\cite{hamdan2024carformer}, conducted experiments in the CARLA simulator~\cite{dosovitskiy2017carla} using object-centered representations. With the introduction of the nuPlan benchmark~\cite{caesar2021nuplan}, subsequent studies~\cite{pdm,hu2023imitation,gameformer,chekroun2023mbappe,cheng2024rethinking,cheng2024pluto} have leveraged this dataset for comprehensive evaluations. These methods encode vectorized observations using scene encoders like PointNet~\cite{qi2017pointnet} or BERT~\cite{devlin2018bert}, as well as motion forecasting models~\cite{nayakanti2023wayformer,cheng2023forecast}. 
However, their effectiveness in out-of-distribution scenarios is suboptimal, \emph{e.g.,} in \textit{close-loop} simulation.

\section{Problem Formulation}

At time step $t$, the objective for mid-to-end autonomous driving is to estimate actions $a_t$ based on the current post-perception results $O_{t}$. The training objective is $P(a_t|O_{t}).$

In autonomous driving, a world model is to predict future states based on actions and observations. The states can be observations (world model) or extracted latent features (latent world model), here we focus on the latter one. Specifically, let $\mathbf{s}_{t}$ be latent features at time step $t$, $O_{1:t}$ and $\hat{a}_{1:t}$ be observations and ground truth actions from time step $1$ to $t$, the training objective for latent world model is $P({\mathbf{s}}_{t+1}|O_{1:t}, \hat{a}_{1:t}).$

We can derive planning-oriented training objective with latent world model accordingly as 
\begin{equation}
    \resizebox{1\linewidth}{!}{$
        P(a_t,{\mathbf{s}}_{t+1} | O_{1:t}, \hat{a}_{1:t-1})=P(a_t|{\mathbf{s}}_{t+1},O_{1:t}, \hat{a}_{1:t-1})P({\mathbf{s}}_{t+1}|O_{1:t}, \hat{a}_{1:t-1}).
    $}
\end{equation}
We omit $\hat{a}_{t}$ here as it is the variant needed to estimate. The planner should react based on the current observation $O_{t}$, but including historical actions $\hat{a}_{1:t-1}$ can lead to a ‘copycat’ phenomenon~\cite{ortega2021shaking,cheng2024rethinking}. 

To address these issues, we propose using the estimation of actions for the world model instead of the real one. By introducing an intermediate estimation term ${a}'_{1:t}$, the planning-oriented objective becomes
\begin{equation}
    \begin{split}
        & P(a_t,\mathbf{s}_{t+1}, {a}'_{1:t} | O_{1:t}) \\
        &= P(a_t,\mathbf{s}_{t+1} | O_{1:t}, {a}'_{1:t})P({a}'_{1:t}|O_{1:t})   \\
        &= \underbrace{P(a_t|\mathbf{s}_{t+1},O_{1:t},{a}'_{1:t})P({a}'_{1:t}|O_{1:t})}_\text{multiple probabilistic planner} \underbrace{P(\mathbf{s}_{t+1}|O_{1:t}, {a}'_{1:t}).}_\text{world model}
    \end{split}
    \label{eq.overall}
\end{equation}
We formulate the action using a Gaussian Mixture Model (GMM), denoted as $\bar{A}$ and the latent state as a Gaussian distribution, denoted as $\bar{\mathbf{s}}$.
\section{Methods}
\begin{figure*}
    \centering
    \includegraphics[width=\linewidth]{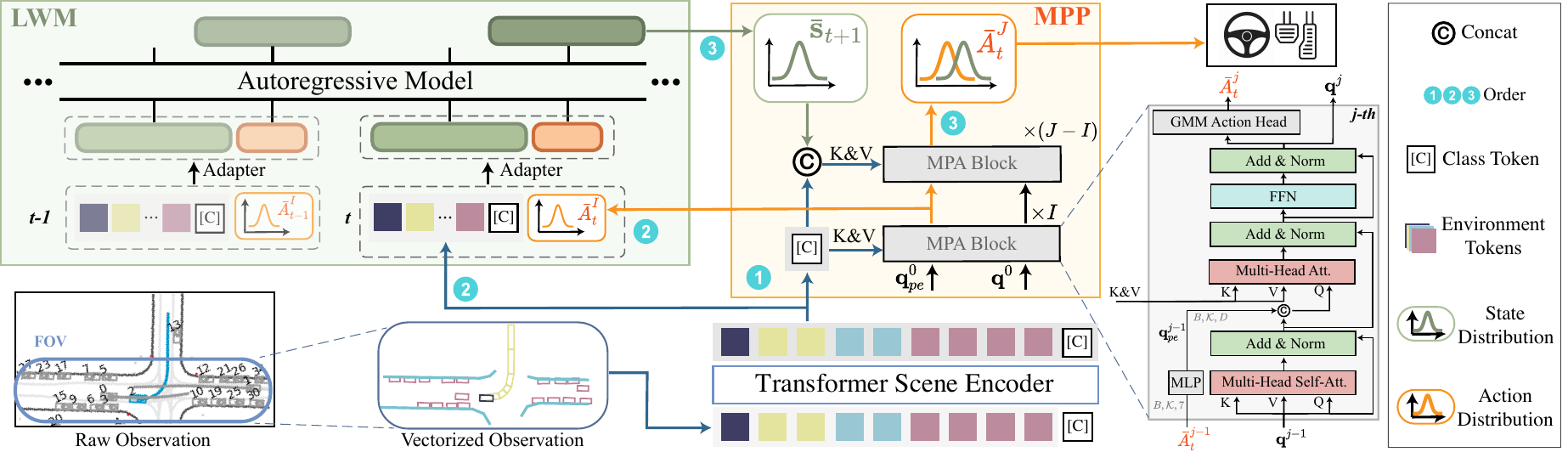}
    \caption{Overall pipeline for \modelname{}. The scheme is in three steps. The class token from scene encoder is first fed into a Multiple Probabilistic Planner (MPP) which will generate an intermediate action distribution \textcolor{actioncolor}{$\bar{A}^I_{1:t}$} from its $I$ layer. Then the Latent World Model (LWM) is introduced to generate latent state distribution \textcolor{statecolor}{$\bar{\mathbf{s}}_{t+1}$} based on $\mathbf{h}_{1:t}$ and \textcolor{actioncolor}{$\bar{A}^I_{1:t}$}. Lastly, the final execution signal is generated by the $J$ layer output from planner aid by \textcolor{statecolor}{$\bar{\mathbf{s}}_{t+1}$}.}
    \label{fig.overall}
    \vspace{-5mm}
\end{figure*}

To realize Eqn.~\ref{eq.overall}, we propose \modelname{}, as illustrated in Fig.~\ref{fig.overall}. The raw observation is first vectorized and then fed to a scene encoder.
The intermediate action distribution is generated by an intermediate layer of the MPP. Upon receiving the intermediate action, the LWM predicts the next latent state and formulates it as a distribution. The action distribution and latent state distribution are then combined through subsequent layers of the MPP, resulting in a mixture distribution from which the final control signal is derived.

\subsection{Input Representation and Context Encoding}
\label{sec.input}

At each time step, the raw observation is vectorized using object-centered representation as described in~\cite{renz2022plant}, resulting in $O_{t} \in \mathbb{R}^{N \times 6}$, where $N$ represents the number of segments. The attributes of each segment are consistent with the definitions provided in~\cite{renz2022plant}.
We only consider context that within the Field of View (FOV) under the ego vehicle coordinate system, with a certain width $w_f$ and height $h_f$. To extract this object-centered vectorized scene information, we utilize BERT~\cite{devlin2018bert} as our scene encoder. Given a sequence of observations $O_{1:t}$, the collection of class token and environment tokens $\mathbf{h}_{1:t} \in \mathbb{R}^{N \times D}$ is obtained by feeding $O_{1:t}$ through BERT,  where $D$ is the dimension of BERT.


\subsection{World Model for Latent Prediction}
\label{sec.worldmodel}

We formulate the latent state prediction as a next token prediction task using autoregressive model as shown in Fig.~\ref{fig.overall}. We refer latent state as latent state token here.
The Latent World Model (LWM) is designed to predict the next latent state token using action tokens and previous latent state tokens (both generated by an adapter that takes $\bar{A}$ and $\mathbf{h}$).

\subsubsection{Adapter}
The adapter generates two types of tokens: action tokens and latent state tokens. For the action tokens, given actions, each dimension of input actions is independently mapped into a $D$-dimensional space via a linear layer. Consequently, considering waypoints as the action space, the action tokens at time step $t$ is represented as $\mathbf{a}_{t}=(\mathbf{a}_{t,x},\mathbf{a}_{t,y},\mathbf{a}_{t,yaw}) \in \mathbb{R}^{3\times D}$.
It is important to note that during training, the input actions for LWM is estimated from the planner, whereas during inference, it is derived from the actual executed historical action sequence. For the latent state tokens, we employ several stacked standard transformer cross-attention blocks that use $M$ learnable queries $(M < N)$ to encode $\mathbf{h}_{1:t} \in \mathbb{R}^{N \times D}$ into latent space~\cite{li2023blip}, followed by a distribution head $\phi$ to parameterize it as a Gaussian distribution $\hat{\mathbf{s}}_t \sim \mathcal{N}((\mu_{\phi}\circ\operatorname{CrossAtt})(\mathbf{h}_t),(\sigma_{\phi}\circ\operatorname{CrossAtt})(\mathbf{h}_t))$, where $\mu_{\phi}$ and $\sigma_{\phi}$ are multilayer perceptrons (MLPs). Thus, the latent state token (or latent state) is sampled from the latent state's distribution, $\mathbf{s}_t = \operatorname{sample}(\hat{\mathbf{s}}_t), \mathbf{s}_t \in \mathbb{R}^{M\times D}.$

\subsubsection{Latent world model}

At each time step, input tokens are ordered as `action - observation'. The input to an autoregressive transformer~\cite{radford2019language} is expressed as $(\mathbf{a}_1, \mathbf{s}_1, \ldots, \mathbf{a}_t, \mathbf{s}_t)$.
We utilize a factorized spatio-temporal position embedding to encode token positions. 
The output latent state token $\mathbf{s}_{t+1}'$ is modeled as a Gaussian distribution via a distribution head $\theta$. Therefore the next latent state is sampled from distribution using $\mathbf{s}_{t+1} = \operatorname{sample}(\bar{\mathbf{s}}_{t+1} \sim \mathcal{N}(\mu_\theta(\mathbf{s}_{t+1}'), \sigma_\theta(\mathbf{s}_{t+1}'))).$

\subsection{Multiple Probabilistic Planner}
\label{sec.planner}

The multiple probabilistic planner is composed of Multiple Probabilistic Action (MPA) blocks, as shown in Fig.~\ref{fig.overall}. Each layer will generate a action distribution and the ultimate control action for ego vehicle is provided by the action distribution of the last layer.

\subsubsection{Multiple probabilistic modeling for decision-making}

Considering waypoints as the action space, the ground truth actions are denoted as $\hat{a}=[\hat{a}_x, \hat{a}_y, \hat{a}_{yaw}] \in \mathbb{R}^{3}$. Prior researches~\cite{shi2022motion,chai2019multipath} used Gaussian Mixture Models (GMM) for trajectory prediction, focusing on $x, y$. Recognizing the spatial relationship between ${a}_{yaw}$ and its horizontal actions ${a}_{x}, {a}_{y}$, we propose modeling ${a}_{x}, {a}_{y}$ with a Gaussian mixture distribution and ${a}_{yaw}$ with a Laplace distribution. 

Specifically, we predict the probability $p$ and parameters $(\mu_x, \mu_y, \sigma_x, \sigma_y, \rho)$ for each Gaussian component follows $G^j=\text{MLP}(\mathbf{q}^j),$
where $G^j \in \mathbb{R}^{\mathcal{K}\times6}$ represents the parameters of $\mathcal{K}$ Gaussian components $\mathcal{N}_{1:\mathcal{K}}(\mu_x, \sigma_x; \mu_y, \sigma_y; \rho)$ and their associated probabilities $p_{1:\mathcal{K}}$. The query content from the $j$-th layer is denoted as $\mathbf{q}^j \in \mathbb{R}^{\mathcal{K}\times D}$. 

For $a_{yaw}$, we use another network predicts its parameter $\mu_{yaw}$, $L^j = \text{MLP}(\mathbf{q}^j)$,
where $L^j \in \mathbb{R}^{\mathcal{K}\times1}$ is the parameter for each Laplace component $\operatorname{Laplace}_{1:\mathcal{K}}(\mu_{yaw},1)$. The final action distribution in the $j$-layer is $\bar{A}^j=[G^j,L^j] \in \mathbb{R}^{\mathcal{K}\times7}$,
where $[\cdot,\cdot]$ denotes concatenation. The actions is sample from mixture distribution by using the expectation of the Gaussian component with the highest probability, as well as the Laplace component following
\begin{equation}
    \begin{split}
        & k^* = \underset{k \in (1,\mathcal{K})}{\arg \max}~p, \\
        & a^j = \operatorname{sample}(\bar{A}^j) = (\mathbb{E}(\mathcal{N}_{k^*}), \mathbb{E}(\operatorname{Laplace}_{k^*})).
    \end{split}
\end{equation}
\subsubsection{MPA block}
The detail design of MPA block is shown in Fig.~\ref{fig.overall}. Given action distribution $\bar{A}^{j-1}$ and query content features $\mathbf{q}^{j-1}$ from the previous layer, the query content feature is updated via a self-attention module. The action distribution is embedded into a token through an MLP, serving as a query position embedding for the cross-attention module to extract features from the latent world model and scene encoder outputs. The query content features and query position embedding are concatenated following practices in~\cite{meng2021conditional,shi2022motion}.
For $j=1$, where $\bar{A}^{0}$ is unavailable, we initialize $\mathbf{q}_{pe}^{0} \in \mathbb{R}^{\mathcal{K}\times D}$ as a learnable token. $\mathbf{q}^0$ is initialized with zero following previous practices~\cite{detr}. For $j \le I$ where $\bar{\mathbf{s}}_{t+1}$  is lack, we omit it in the key and value.

\subsection{Loss Function}
\subsubsection{World model}
Given $O_{1:t}$ and $a'_{1:t}$,
the training objective for the world model is to minimize the Kullback-Leibler (KL) divergence between the adapter's output $\hat{\mathbf{s}}_{2:t}$ and the estimated latent state distribution $\bar{\mathbf{s}}_{2:t}$
\begin{equation}
    \mathcal{L}_{world} = -\sum_{i=2}^t D_{KL}(\hat{\mathbf{s}}_{i} \Vert \bar{\mathbf{s}}_{i}).
\end{equation}
\subsubsection{Planner}
The predicted distribution for the ego vehicle's actions is formulated as
\begin{equation}
    \resizebox{\linewidth}{!}{
    $
        \sum_{k=1}^\mathcal{K}
            p_k 
            \cdot 
            \mathcal{N}_{k}(\hat{a}_x-\mu_x, \sigma_x; \hat{a}_y-\mu_y, \sigma_y; \rho) 
            \cdot
            \operatorname{Laplace}_{k}(\hat{a}_{yaw}-\mu_{yaw}, 1).
    $}
    \label{eq.obj_gmm}
\end{equation}
We use negative log-likelihood loss to maximize the likelihood of the ego vehicle's ground truth action $\hat{a}$. Thus, the loss function for action is formulated as
\begin{align}
    \nonumber
   \mathcal{L}_{gmm} & = - \log(p_s) -\log \mathcal{N}_s(d_x, \sigma_x; d_y, \sigma_y; \rho) \\
   &- \log \operatorname{Laplace}_{s}(d_{yaw}, 1),
   \label{eq.gmmloss}
\end{align}
where $d_x=\hat{a}_x-\mu_x$, $d_y=\hat{a}_y-\mu_y$, $d_{yaw}=\hat{a}_{yaw}-\mu_{yaw}$, and $\mathcal{N}_s$ refers to the selected positive component for optimization, as do $\operatorname{Laplace}_{s}$ and $p_s$. The selection procedure will be discussed in Section~\ref{sec.exp}. 
Since each MPA block contains a GMM action head, the final loss is the average of Eqn.~\ref{eq.gmmloss} across all decoder layers. The final loss for \modelname{} is formulated as 
\begin{equation}
    \mathcal{L} = 0.001 \times \mathcal{L}_{world} + \mathcal{L}_{gmm}.
\end{equation}

\section{Experiments}
\label{sec.exp}
\subsection{Environment and Dataset}
All experiments are conducted using the recently released simulator Waymax~\cite{gulino2024waymax} driven by WOMD dataset (v1.1.0)~\cite{womd}.
Each scenario has a sequence length of 8 seconds, recorded at 10 Hz. Agents are controlled at a frequency of 10 Hz, with a maximum of 128 agents per scenario. The training set comprises 487,002 scenarios, while the validation set includes 44,096 scenarios. 
The simulation will not be terminated until it reaches the maximum length of 8 seconds.

\begin{table*}[]
    \centering
    \caption{Comparison with state-of-art methods. Non-ego agents are controlled by IDM~\cite{idm}. `LT' and `DF' under \textbf{Route} shorts for Logged Trajectory and Drivable Futures (unavailable),  respectively.}
    \vspace{-3mm}
    \label{tb.sota}
    \resizebox{\linewidth}{!}
    {
        \begin{tabular}{l|cc|ccccc}
            \whline{0.7pt}
            \textbf{Methods} &
              \textbf{Route} &
              \textbf{Action Space} &
              \textbf{mAR@{[}95:75{]}$\uparrow$} &
              \textbf{AR@{[}95:75{]}$\uparrow$} &
              \textbf{OR$\downarrow$} &
              \textbf{CR$\downarrow$} &
              \textbf{PR$^*\uparrow$} \\ \whline{0.7pt}
            Stationary Agent  & -     & -                         & 20.31      & 26.3      & 0.29        & 0.63       & 35.29                               \\
            Logged Oracle     & -     & -                         & 97.01      & 96.14     & 0.63        & 3.27       & 100                                 \\ \hline 
            Waymax-BC~\cite{gulino2024waymax}         & LT+DF &                           & -          & -         & 13.59±12.71 & 11.20±5.34 & {\color[HTML]{9B9B9B} 137.11±33.78} \\
            EasyChauffeur-PPO~\cite{xiao2024easychauffeur} & LT    & \multirow{-2}{*}{Bicycle} & \underline{78.716}     & \underline{88.66}     & 3.95        & 4.72       & \underline{98.26}                          \\ \hline
            Wayformer~\cite{nayakanti2023wayformer}         & LT+DF &                           & -          & -         & 7.89        & 10.68      & {\color[HTML]{9B9B9B} 123.58}       \\
            Waymax-BC~\cite{gulino2024waymax}         & LT+DF &                           & -          & -         & 4.14±2.04   & 5.83±1.09  & {\color[HTML]{9B9B9B} 79.58±24.98}  \\
            PlanT$^\dag$~\cite{renz2022plant}            & LT    &                           & 77.79±2.12 & 87.41±0.4 & \textbf{1.9±0.34}    & \textbf{2.87±0.18}  & 95.76±1.03                          \\ 
            \rowcolor{baselinecolor}
            \modelname{} (Ours) &
              LT &
              \multirow{-4}{*}{Waypoints} &
              \textbf{89.3±0.83} &
              \textbf{94.07±0.21} &
              2.33±0.13 &
              3.17±0.04 &
              \textbf{99.57±0.1} \\ \hline
            \end{tabular}
    }
    \vspace{-5mm}
    \end{table*}

\subsection{Scene Categorization}
\label{sec.sceneclass}
\begin{figure}
    \centering
    \includegraphics[width=1\linewidth]{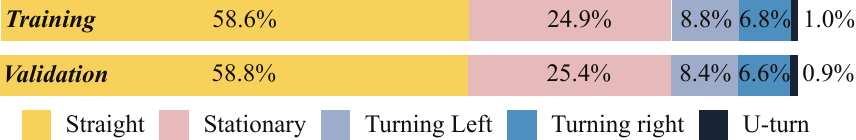}
    \caption{The percentages of the episode number for each driving scenario in the training and validation sets.}
    \label{fig.percentage}
    \vspace{-5mm}
\end{figure}

We found the original metrics in Waymax~\cite{gulino2024waymax} can not reflects the long tail problem due the lack of scenario type information. Therefore, we categorize the driving scenarios into five representative types: Stationary, Straight, Turning Left, Turning Right, and U-turn. The categorization is based on the pattern of the ego vehicle’s expert trajectory, which reflects its intention. Specifically, given the route’s maximum curvature $\kappa$ and the heading difference $\delta$ between the starting and ending locations, we classify Straight, Turning, and U-turn scenarios as follows
\begin{equation}
\resizebox{\linewidth}{!}{$
    \text{Scene} =
\begin{cases}
\text{Turning} & \text{if } \left( 0.03 < \kappa < 0.18 \text{ and } \delta > 0.2 \right) \text{ or } \left( 0.1 < \kappa < 0.18 \right), \\
\text{U-turn} & \text{if } \kappa \geq 0.18, \\
\text{Straight} & \text{otherwise}.
\end{cases}
$}
\end{equation}
All the thresholds have been tuned empirically.
The proportion of each driving scenario is visualized in Fig.~\ref{fig.percentage}. The proportion of each type is approximately the same in both the training and validation sets.
The Straight scenarios constitutes the majority of episodes at 59\%, followed by the Stationary scenarios at 25\%. The number of Turning episodes for the right and left is almost equal, while U-turns appear in only 1\% of the entire set. Based on this statistical result, the original evaluation metric can be insufficient, as the overall performance can be easily dominated by the Straight scenarios, thereby neglecting Turning and U-turn scenarios, which are equally vital. To address this, we propose a new metric, mean Arrival Rate (mAR), which will be introduced in next section.

\subsection{Evaluation Metrics}
\label{sec.evaluation}
All experiments are conducted under \textit{close-loop} evaluation following~\cite{gulino2024waymax}. The \textbf{Off-road Rate (OR)} and \textbf{Collision Rate (CR)} follows the same description in~\cite{gulino2024waymax}. The definition for \textbf{Progress Ratio (PR)} is consistent while the maximum value in this paper is 100\%, as the future derivable area is not disclosed in WOMD. Other metrics we used is detailed as: 
\begin{itemize}
	\item \textbf{Arrival Rate under $\tau \%$ (AR@$\tau$).} It determines if the ego vehicle has traveled $\tau \%$ of the route safely. For example, AR@50 refers to the safe travel of 50\% of the route. To avoid the metric dominated by PR, we report AR@[95:75] to compare different algorithms' performance,
        \begin{equation}
            \nonumber
            \text{AR@[95:75]} = (\text{AR@95} + \text{AR@90} + \cdots +\text{AR@75}) / 5.
        \end{equation}
    \item \textbf{Mean Arrival Rate (mAR).} It represents the average AR across all categorized scenarios, 
    \begin{equation}
    \nonumber
        \text{mAR} = (\text{AR}_\text{Straight} + \cdots + \text{AR}_\text{U-turn}) / 5.
    \end{equation}
\end{itemize}

\subsection{Implementation Details}
\subsubsection{Hyperparameters} The FOV dimensions are set to $w_f=80$ and $h_f=20m$. The time step $t$ is set to 2. For the scene encoder, we use a randomly initialized BERT-mini model with feature dimension $D$ of 256. In the MPP, we use $I=1$, the 1st layer to estimate the intermediate distribution and $J=3$ for the final action prediction. 
The mode $\mathcal{K}$ of GMM is set to 6 following previous practice in trajectory prediction~\cite{shi2022motion}. 
The LWM's cross-attention module has 4 layers, 4 heads per layer, and employs $M=32$ learnable queries.
For the autoregressive model, we use a randomly initialized GPT-2~\cite{radford2019language} model with 8 layers and 8 heads for each layer. The encoder is pretrained using our re-implementation of PlanT\cite{renz2022plant} for faster convergence. We set the batch size to 2,500 and use the Adam optimizer. The learning rate is initialized at $2 \times 10^{-4}$ and decays to 0 over 10 epochs using a cosine scheduler. 

\subsubsection{Positive actions selection} The procedure for positive action selection is analogous to label assignment in detection, where the ego vehicle’s next location is considered as a rotated bounding box in the top-down view. 
The rotation Intersection over Union (IoU)~\cite{zhou2019iou} between proposed actions and the ground truth actions are calculated. A proposal is considered positive if it meets one of the following criteria: (1) It has the largest IoU; (2) Its IoU is greater than 0.7. Conversely, a proposal is considered negative if its IoU is less than 0.3. Proposals with IoU values in between these thresholds are not assigned any label.

\subsection{Performance Comparison}
The comparative analysis with other methods on Waymax~\cite{gulino2024waymax} is presented in Table~\ref{tb.sota}. Non-ego agents are controlled by the Intelligent Driver Model (IDM)~\cite{idm}, as implemented in previous works~\cite{gulino2024waymax,xiao2024easychauffeur}. The asterisk ($^*$) on PR indicates that this metric cannot be fairly compared under some methods due to the absence of drivable future under `Route'. This discrepancy is highlighted in \textcolor{gray}{gray}. The mean and standard deviation are reported under three random seeds.

The first two rows represent scenarios where the ego vehicle is stationary and controlled by actions from the driving log, respectively, indicating the lower and upper boundaries of the benchmark. 
The dagger ($^\dag$) on PlanT~\cite{renz2022plant} denotes our re-implementation to fit the dynamic space in this benchmark.
Focusing solely on OR and CR is insufficient, as the risk is proportional to travel distance. Therefore, mAR and AR are more indicative of overall performance. Our method achieves the best results in mAR, AR, and PR across all approaches, demonstrating superior performance. Notably, our method's AR is only 2\% lower than the Logged Oracle, indicating expert-level performance. However, the Logged Oracle's mAR is 8\% higher than ours, suggesting that certain scenarios remain challenging. EasyChauffeur ranks second after our model in mAR and AR, followed closely by PlanT. While PlanT performs best in OR and CR, it has one of the shortest travel distances except Waymax-BC. Both PlanT and EasyChauffeur have mAR below 80\%, indicating weaker performance in some long-tail scenarios.

\subsection{Ablation Studies}
In ablation studies, all non-ego agents is controlled using actions from expert demonstration for its lower evaluation time than IDM.
\subsubsection{Overall ablation results}

The overall ablation results under type-invariant metrics are presented in Table~\ref{tb.overall-ab}. Here, LWM and MPP denote the Latent World Model and Multiple Probabilistic Planner, respectively. The first row, which omits both is baseline. We take PlanT as baseline to compare with since it utilizes the same encoder as ours. 
The second row with the lowest AR indicates that naively using LWM for decision-making results in self-delusion.

Employing only MPP achieves results similar to PlanT, suggesting that multi-probabilistic action modeling alone provides limited performance improvement. However, incorporating both LWM and MPP results in a significant performance boost across all metrics, except a slight deterioration in the off-road rate. This phenomenon demonstrates that the performance gains originate from the combination of both components.
\begin{table}[]
\centering
\caption{Overall ablation studies under type invariant metric.}
\label{tb.overall-ab}
\begin{tabular}{cc|cccc}
\whline{0.7pt}
\textbf{LWM} & \textbf{MPP} & \textbf{AR@[95:75]}$\uparrow$  & \textbf{OR}$\downarrow$  & \textbf{CR}$\downarrow$  & \textbf{PR}$\uparrow$ \\ \whline{0.7pt}
 &  & 87.39 & \textbf{2.15} & 3.8 & 95.11 \\
\checkmark &  & 54.46 & 2.35 & 17.93 & 84.56 \\
 & \checkmark & 88.55 & 2.65 & 3.17 & 97.09 \\
\checkmark & \checkmark & \textbf{94.82} & 2.58 & \textbf{2.31} & \textbf{99.55} \\ \hline
\end{tabular}
\end{table}

\begin{table}[]
    \centering
    \caption{Per-type results on ablation studies.}
    \label{tb.perclass-ab}
    \resizebox{\linewidth}{!}{
        \begin{tabular}{c|c|cccc}
            \whline{0.7pt}
            \textbf{Types}                                                                          & \textbf{Model} & \textbf{AR@[95:75]}$\uparrow$ & \textbf{OR}$\downarrow$   & \textbf{CR}$\downarrow$  & \textbf{PR}$\uparrow$ \\ \whline{0.7pt}
            \multirow{3}{*}{\begin{tabular}[c]{@{}c@{}}Easy\\ (Stationary/\\Straight)\end{tabular}} & PlanT          & 92.605              & \textbf{1.09} & 2.74        & 97.265      \\
                                 & MPP     & 92.93           & 1.36           & 2.38           & 98.31           \\
                                 & MPP+LWM & \textbf{96.895} & 1.21           & \textbf{1.8}   & \textbf{99.815} \\ \hline
            \multirow{3}{*}{\begin{tabular}[c]{@{}c@{}}Medium \\ (Turning)\end{tabular}}           & PlanT          & 70.665              & 6.125         & 7.99        & 89.565      \\
                                 & MPP     & 74.74           & 7.24           & 6.58           & 93.78           \\
                                 & MPP+LWM & \textbf{89.655} & \textbf{5.93}  & \textbf{3.625} & \textbf{99.295} \\ \hline
            \multirow{3}{*}{\begin{tabular}[c]{@{}c@{}}Hard\\ (U-turn)\end{tabular}}               & PlanT          & 50.16               & 23.94         & 6.38        & 87.56       \\
                                 & MPP     & 49.15           & 38.03          & 7.71           & 95              \\
                                 & MPP+LWM & \textbf{78.78}  & \textbf{17.29} & \textbf{3.46}  & \textbf{99.24}  \\ \hline
            \multirow{3}{*}{All} & PlanT   & 75.34           & 7.674          & 5.568          & 92.244          \\
                                 & MPP     & 80.28           & 8.334          & 4.416          & 96.208          \\
                                 & MPP+LWM & \textbf{90.376} & \textbf{6.314} & \textbf{2.862} & \textbf{99.492} \\ \hline
            \end{tabular}
    }
    \vspace{-5mm}
\end{table}

\newcommand{\addFig}[1]{\includegraphics[width=0.24\linewidth]{pics/gallery/#1.pdf}}
\begin{figure*}
  \centering
  \footnotesize
  \setlength\tabcolsep{0.4mm}
  \renewcommand\arraystretch{0.8}
  \begin{tabular}{ccccc}
    \multicolumn{5}{c}{\includegraphics[width=.92\linewidth]{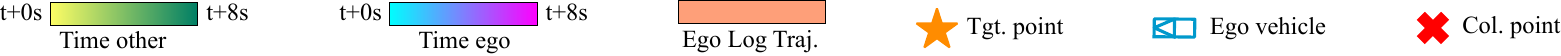}} \\
    \rotatebox{90}{~~~~~~Straight} & 
    \addFig{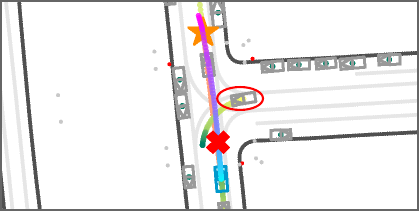} & \addFig{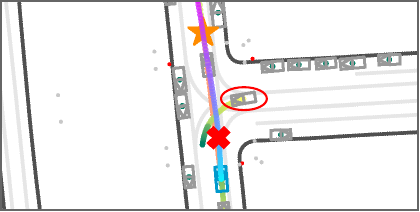} & \addFig{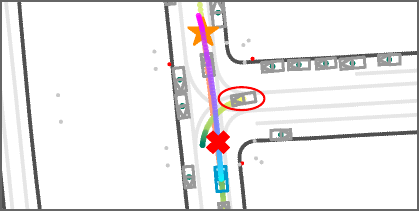} & \addFig{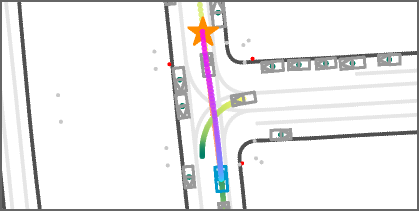} \\
    \rotatebox{90}{~~~Turning Right} & 
    \addFig{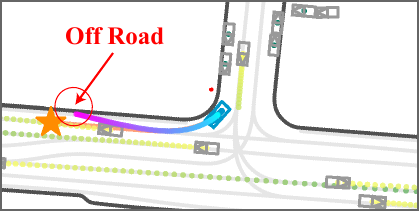} & \addFig{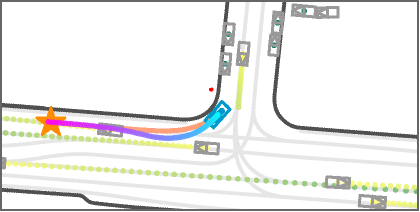} & \addFig{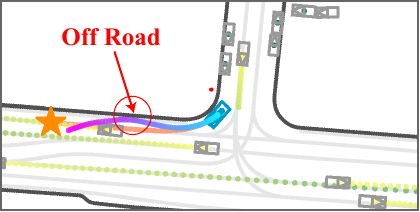} & \addFig{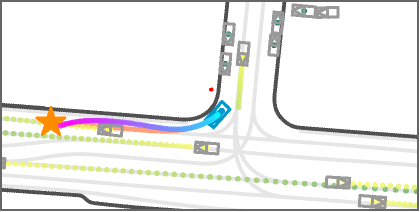} \\
    \rotatebox{90}{~~~Turning Left} & 
    \addFig{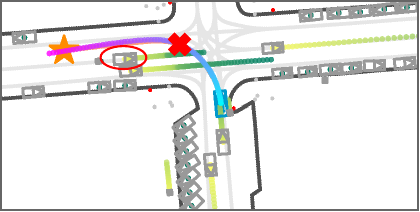} & \addFig{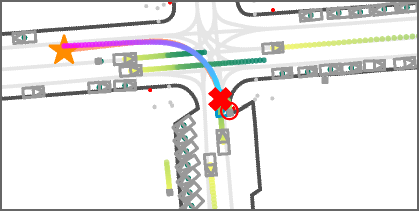} & \addFig{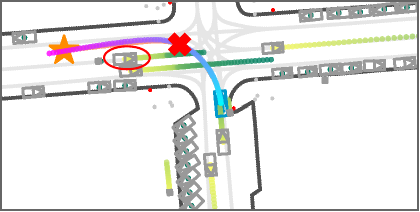} & \addFig{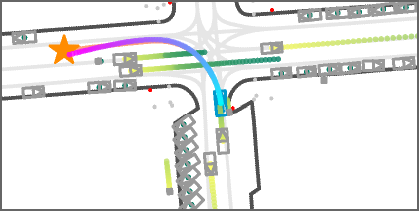} \\
    \rotatebox{90}{~~~~~~~U-turn} & 
    \addFig{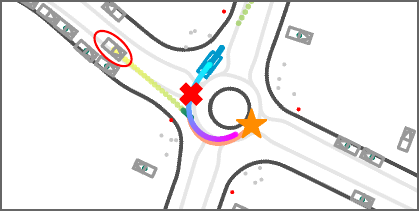} & \addFig{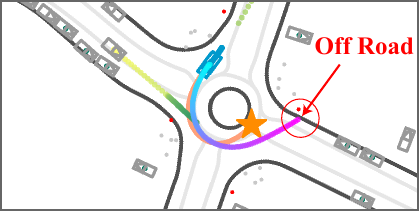} & \addFig{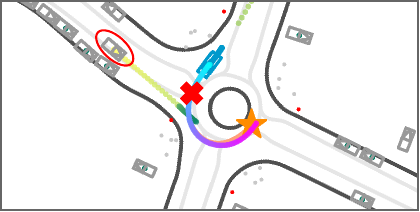} & \addFig{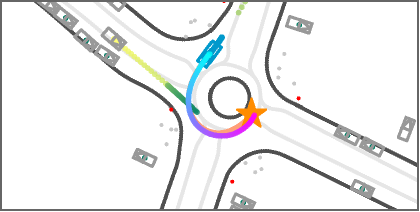} \\
    & (a) PlanT & (b) EasyChauffeur-PPO & (c) GMM baseline & (d) \modelname{} (Ours) \\
  \end{tabular}
  \caption{Visualization results of \modelname{} against other three methods in four driving scenarios. For a detailed explanation, please refer to Section~\ref{sec.vis}.}
  \vspace{-2mm}
  \label{fig.vis}
\end{figure*}

\subsubsection{Ablation results under per-type metrics}
In addition to the type-invariant metrics in Table~\ref{tb.overall-ab}, we also provide per-type results in Table~\ref{tb.perclass-ab}. Based on scenario types, the results are grouped under four tags reflecting their difficulty: Easy, Medium, Hard, and All.

The performance of MPP and PlanT is nearly identical on Hard and Easy scenarios while MPP performs better on Medium scenarios. In the Easy scenario, AR only increased from approximately $92\%$ to $97\%$ for the incorporation of LWM, while the performance in Medium and Hard scenarios experienced a significant improvement. This demonstrates that the incorporation of the Latent World Model enables the planner to learn more complex dynamic relationships and utilize such high-context information for better decision-making in challenging driving scenarios.

\begin{table}[]
    \centering
    \caption{Necessity of multi-probabilistic hypotheses.}
    \label{tb.gmm}
    \resizebox{\linewidth}{!}{
    \begin{tabular}{c|cc}
    \whline{0.7pt}
    \textbf{Methods} & \textbf{mAR@[95:75]}$\uparrow$ & \textbf{AR@[95:75]}$\uparrow$ \\ \whline{0.7pt}
    GMM baseline & 80.39 & 87.42 \\
    + LWM \emph{w.} estimated action & 88.94 (+8.55) & 95.27 (+7.85) \\ \hline
    PlanT & 75.34 & 87.39 \\
    + LWM \emph{w.} estimated action & 81.13 (+5.79) & 89.77 (+2.38) \\ \hline
    \end{tabular}
    }
    \vspace{-5mm}
    \end{table}
    
\subsubsection{Necessity of multi-probabilistic hypotheses}

We provide ablation studies in Table~\ref{tb.gmm} to investigate the necessity of multi-probabilistic modeling when incorporating the LWM. The first row, labeled ‘GMM baseline,’ represents the use of a single layer of the MPP to predict actions and no world model knowledge is incorporated. The second row corresponds to using two layers of the MPP for action prediction. The first layer is for intermediate action estimation and the second is to incorporate LWM's output.
Similarly, for ‘PlanT+LWM,’ we employ another MLP to fuse world model's output. It is noteworthy that actions for the LWM are estimated during training, thereby preventing self-delusion. When comparing the two baselines, the performance on AR is similar, while the GMM baseline demonstrates an advantage in solving some hard scenarios, as evidenced by the mAR metric. 
The performance gain for the integration of ‘GMM+LWM’ is significant, showing improvements of $8.55\%$ and $7.85\%$ for mAR and AR, respectively. On the other hand, the benefits brought by LWM under PlanT are marginal but still outperform the naive GMM baseline.


\subsection{Visualization Results}
\label{sec.vis}
We visualize the behaviors of four different methods across four typical and distinct driving scenarios in Fig.~\ref{fig.vis}. In the straight scenario, other three methods collide with a turning vehicle due to hesitation at the intersection. In contrast, \modelname{} successfully navigates through by making a decisive decision. A similar outcome is observed in the U-turn scenario, where both PlanT and the GMM baseline fail, while EasyChauffeur manages the turn but ultimately goes off the road. During an unprotected left turn, PlanT and the GMM baseline fail to avoid contact with an oncoming vehicle. Although both EasyChauffeur and \modelname{} handle this situation better, EasyChauffeur collides with a pedestrian at the beginning. For the right turn, while all methods exhibit trajectory fluctuations, only EasyChauffeur and \modelname{} complete the turn safely, with the other two going off-road.

\section{Conclusion}
In this paper, we propose \modelname{} to address the challenges of inadequate uncertainty modeling and self-delusion in autoregressive world model-enhanced planners. Our approach represents the environment’s next states and the ego vehicle’s next actions as a mixture distribution, which forms the basis for selecting the planner's final action. The LWM module predicts the distribution of the environment's next state, while the MPP module refines the ego vehicle’s action using LWM's output. \modelname{} outperforms current state-of-the-art methods on the Waymax benchmark and achieves expert-level performance.
\newpage
\bibliographystyle{IEEEtran}
\bibliography{root}

\end{document}